\documentclass[letterpaper, 10 pt, conference]{ieeeconf}  

\IEEEoverridecommandlockouts                              

\usepackage{array}
\usepackage[dvipsnames]{xcolor}
\usepackage{graphics} 
\usepackage{epsfig} 
\usepackage{subfigure}
\usepackage{mathptmx} 
\usepackage{times} 
\usepackage{amsmath} 
\usepackage{amssymb}  
\usepackage[citestyle=numeric-comp, sorting=none]{biblatex}
\usepackage{csquotes}
\usepackage{algorithm, algpseudocode}
\usepackage{hyperref}
\usepackage{todonotes}

\bibliography{references}

\title{\LARGE \bf
Neural Potential Field for Obstacle-Aware Local Motion Planning
}

\author{Muhammad Alhaddad$^{1}$, Konstantin Mironov$^{1,2}$, Aleksey Staroverov$^{2,3}$, and Aleksandr Panov$^{1,2}$
\thanks{$^{1}$Muhammad Alhaddad, Konstantin Mironov, and Aleksandr Panov are with Center of Cognitive Modeling, Moscow Institute of Physics and Technology, Dolgoprudny, 141701, Russia
        {\tt\small mironov.kv@mipt.ru}}%
\thanks{$^{2}$Konstantin Mironov, Aleksey Staroverov and Aleksandr Panov are also with the Artificial Intelligence Research Institute, Moscow, 105064, Russia}%
\thanks{$^{3}$Aleksey Staroverov is also with the Federal Research Center ``Computer Science and Control," Moscow, 117312, Russia}
}

\begin{document}

\maketitle
\thispagestyle{empty}
\pagestyle{empty}

\begin{abstract}
Model predictive control (MPC) may provide local motion planning for mobile robotic platforms. The challenging aspect is the analytic representation of collision cost for the case when both the obstacle map and robot footprint are arbitrary. We propose a Neural Potential Field: a neural network model that returns a differentiable collision cost based on robot pose, obstacle map, and robot footprint. The differentiability of our model allows its usage within the MPC solver. It is computationally hard to solve problems with a very high number of parameters. Therefore, our architecture includes neural image encoders, which transform obstacle maps and robot footprints into embeddings, which reduce problem dimensionality by two orders of magnitude. The reference data for network training are generated based on algorithmic calculation of a signed distance function.  Comparative experiments showed that the proposed approach is comparable with existing local planners: it provides trajectories with outperforming smoothness, comparable path length, and safe distance from obstacles. Experiment on Husky UGV mobile robot showed that our approach allows real-time and safe local planning. The code for our approach is presented at \url{https://github.com/cog-isa/NPField} together with demo video.
\end{abstract}

\section{INTRODUCTION}
Obstacle-aware motion planning is essential for autonomous mobile robots. Various methods may solve this task, including numerical optimization, especially nonlinear Model Predictive Control (MPC) \parencite{sch20a, boj21, zuo20, ji16, zen21, bla11, thi22}. Optimization allows the planner to transform a rough global path into a smooth trajectory, taking into account obstacles and kinodynamic constraints of the robot. 

Obstacle avoidance may be inserted into trajectory optimization either as a set of constraints (e.g., \parencite{sch20a}) or as a penalty term in the cost function (e.g., \parencite{sch14, ji16}). The second approach allow for more flexible trajectory planning via finding a balance between safety and following the reference path; in some cases it may even converge from initial guess that intersect obstacles \parencite{kur22}. However, obstacle representation for this second case is more challenging. On the one hand, collision avoidance in constraint-based optimization consists of detecting the fact of collision. This may be done by projecting the robot's footprint onto the obstacle map. On the other hand, if we use penalty-based optimization, we should define a differentiable penalty function. 
The penalty function forms a repulsive Artificial Potential Field (APF); its gradient directs toward the safer solution \parencite{kha85}. This allows the controller to find a proper balance between the safety of the trajectory and its similarity to the reference path. Therefore, the function which forms the repulsive APF should be differentiable. The \textit{values} of the repulsive APF may be easily computed based on the signed distance function (SDF) from the robot to the nearest obstacle point on the map. However, SDF is computed by specific \textit{algorithms}. It is not a \textit{differentiable function} for the general case. It is easy to define it analytically when two requirements are satisfied: first, the robot is pointwise or circular, and second, the obstacles have known simple geometric shapes. If both the robot footprint and obstacle map are arbitrary, finding accurate and differentiable approximation of the SDF is hard. Simplified versions are used e.g. in \parencite{sch14, sch20b}.

\begin{figure} [t]
\centering
\includegraphics[width=0.48\textwidth]{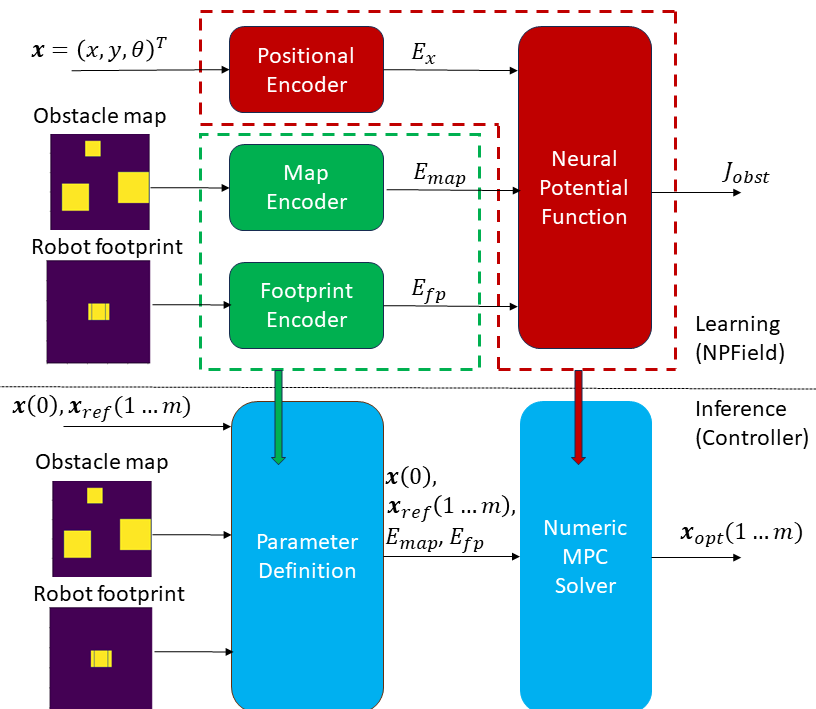}
\caption{Common scheme of the proposed approach. Our controller (bottom half of the figure) consists of a parameter definition module and an MPC solver, which optimize the trajectory for a defined set of parameters.  Our common neural architecture (NPField, top half of the figure) consists of image encoders and a neural potential function (NPFunction). We train NPFunction to predict the obstacle-repulsive potential for a given robot pose and given embeddings of the obstacle map and the robot footprint. The trained neural potential function is used for trajectory optimization within the MPC solver. Map and footprint encoders are removed from the optimization loop to decrease the dimensionality of the MPC problem. They are used for data preparation as both map and footprint are considered constant within the prediction horizon. More precise schemes of this architecture are given in figures \ref{fig:npfield} (Controller) and \ref{fig:net_arc} (Network).}
\label{fig:visual_abstract}
\end{figure} 

We propose a \textit{Neural Potential Field} (NPField) -- a neural network for calculating artificial potential. Our idea is conceptually inspired by the NeRF (Neural Radiance Field) model \parencite{mil20}, which takes the position and orientation of the camera as an input and returns image intensity as an output. Our model takes the position and orientation of the robot together with the obstacle map and robot footprint as input and returns the value of repulsive potential as an output. We aim not to obtain this \textit{values} themselves but to use the trained model within the optimization loop. There are several works where neural networks provide discrete costmaps in which values are used for search-based \parencite{kir23} or sampling-based planning \cite{tri23, cas23}. Our goal instead is to provide \textit{continuously differentiable function}, which gradient is helpful for optimization. The key conceptual scheme of our approach is shown in Fig. \ref{fig:visual_abstract}.

MPC solvers are sensitive to the number of input parameters: a high number of parameters leads to a drastic increase in computational expenses. Map images should not be sent to the solver as they include many cells. We use image encoding within our architecture to reduce the number of parameters by replacing maps and footprints with their more compact embeddings. The top part of the fig. \ref{fig:visual_abstract} presents our neural network architecture, while the bottom part shows the data flow of our controller. The neural network consists of two main parts: encoder block and Neural Potential Function (NPFunction) -- a subnetwork that calculates the potential for a single robot configuration.  Our controller consists of two high-level modules. The first module includes the algorithms, which define parameters for the MPC problem based on actual sensor data. Such a definition is made once per each iteration of the control loop. The second block includes a numerical solver for the control problem, which iteratively optimizes the trajectory based on the pre-defined parameters. NPField is trained as a single architecture and then divided into two parts. Image encoders are inserted into the parameter definition block, while NPFunction is integrated into the numerical MPC solver. For each controller step, encoders are called once, while NPFunction is called and differentiated multiple times within the optimization procedure.

\subsection{Contribution}
This work mainly contributes in the following aspects: 
\begin{itemize}
    \item Novel architecture for MPC local planner, where the neural model estimates collision cost.
    \item Novel neural architecture for calculating APF based on the obstacle map, robot pose, and footprint.
    \item An approach for generating the dataset for training the neural model. 
\end{itemize}
The last subsection of the next section provides a discussion on the place of our approach among the others.

\subsection{Structure}
The rest of the paper is organized as follows. Section II discusses the related works. In section III, we introduce the architecture of our local planner. In section IV, we narrow down to our neural model and describe its architecture and learning. Section V discusses the experiments. Section VI is a conclusion section.

\section{RELATED WORKS}
In this section we fist discuss common approaches to motion planning, then narrow down to collision avoidance in optimization based local planners. After that we discuss existing works, which use neural models within the MPC solvers. Finally we specify the differences of our approach compared to similar works.

\subsection{Planning}
Planning task may be solved by various methods, which could be categorized into the following main groups (see review \parencite{gon16}): search-based planning (most of these methods are based on A* graph search algorithm \parencite{har68}, which is an extension of Dijkstra method \parencite{dij59}), sampling-based planning (most of these methods are based either on Rapidly-exploring Random Trees \parencite{lav01} or on Probabilistic RoadMaps \parencite{kav96}), motion primitives and trajectory optimization. We consider optimization-based planning in this work.

Depending on the statement, we can define two groups of planning tasks -- global path planning (define a reference of intermediate robot configurations based on given initial and destination configuration) and local motion planning (define a smooth trajectory based on a given part of the global plan taking in mind kinodynamic constraints). The artificial potential field was initially proposed \parencite{kha85} for global planning: the robot's path is obtained as a trajectory of the gradient descent in the potential field from the starting point towards the destination point. This planning approach can easily  stuck in the local minimum. Therefore, it is less popular than A*, RRT, or PRM. However, it is still useful \parencite{kim06, ren06, scz22}. To avoid stucking in local minima, trajectory optimization is often done locally together with global planners \parencite{sch20a, thi22}. Global planner generates a rough suboptimal path, which is then optimized. 

Trajectory optimization may be considered in two statements \parencite{sch20a} - holistic computation (made offline before the motion; no strict real-time constraints) and model predictive control (sequential online optimization of near parts of the path during the motion). In the first case, there are no strict limits for the calculation time as well as for the length and complexity of the trajectory. There are some specfifc approaches for this case, such as CHOMP \parencite{rat09}, STOMP \parencite{kal11}, and TrajOpt \parencite{sch14}. CHOMP consider collision avoidance as constraints and therefore require collision-free initial guess. It may work with row representation of obstacles such as Occupancy Grid or Voxel Map for 3D planning tasks. TrajOpt consider collision avoidance as a penalty and may converge from initial guess, which include collision, however, it require obstacles to be represented as polytops. In the second case, the calculation time is limited according to the replanning rate of the system.

\subsection{Obstacle models for trajectory optimization}
Collision detection itself is considered in many works, e.g., \parencite{gil88, zim22, zha22}; we are now interested in analytical models suitable for trajectory optimization. The safe path may be guaranteed using convex approximations of the free space \parencite{sch20a, sch20b}. The disadvantage of such an approximation is that the free space outside the approximated region is prohibited. Alternatively, obstacles may be approximated instead of free space \parencite{ji16, zen21, thi22, bla11}. Interception of the trajectory with the borderlines of the approximated regions may be modeled within the MPC solver. The approaches above require modeling either free space or obstacles as simple geometric shapes, such as points \parencite{ji16}, circles \parencite{sch20a, zen21}, polygons \parencite{bla11, thi22}, or polylines \parencite{zie14}. The question of how to obtain this representation from the common obstacle map is often not considered. Also, some approaches can be used only with discrete-time process models \parencite{thi22}.

In the case when it is impossible or too complicated to provide differentiable collision models, one can use less stable techniques based on numerical gradients \parencite{sch14}, stochastic gradients \parencite{kal11} or gradient-free sampling-based optimization (Model Predictive Path Integral \parencite{wil16, wil17}). We consider another option, where a neural model approximates the repulsive potential. A number of works exist \parencite{daw23} on learning Control Barrier Functions for ensuring the safety of mobile systems such as drones \parencite{ada22} and cars \parencite{abd23} within the controller. The work \parencite{kim22a} provides differentiable collision distance estimation for a 2D manipulator based on a graph neural network. \parencite{kur22} use the loss function of the network as a collision penalty: the trajectory is optimized during the network training for the fixed obstacle map.

\subsection{Neural Models within MPC Optimization}
Integration of neural models into the MPC control loop was considered in a number of tasks. The challenging aspect here is the high computational cost of deep neural models. Accurate deep models by \parencite{pun15, sav22} were not real-time and were presented in simulation as a proof of concept. Real-time inference may be achieved by significantly reducing model capacity \parencite{sav22}. Approaches \parencite{spi21, che22, kim22b} insert lightweight network into realtime MPC control loop. \parencite{sal23} achieve use of the deep neural model within real-time Acados MPC solver \parencite{acados} by introducing ML-CasADI framework \parencite{ml_casadi}. Experiments by \parencite{sal23} showed that direct insertion of the neural model into MPC-solver is effective for the networks with up to 50,000 parameters. Most works above use neural networks for approximating the model of process dynamics. On the contrary, we are interested in approximating obstacle-related cost terms and control barrier functions \parencite{ada22, abd23, daw23}. In \parencite{med23}, a neural network was used to update a cost function for manipulator visual servoing. Its architecture includes a neural encoder for camera images and a cost-update network for quadratic programming. There is also a set of works where a neural network was applied for choosing weighting factors for various terms of the cost function, e.g., \parencite{nov19, wan20}.

\subsection{Place of our approach among the others}
We propose a neural model for estimating repulsive potential, which has the following properties. 1) It provides obstacle avoidance for mobile robotic platforms. 2) It is differentiable. 3) It exploits the capacity of deep neural models with more than 50,000 parameters. 4) It reproduces obstacle maps with complicated, non-convex structures and allows for optimization of long trajectories within this map. 5) It is integrated into the MPC controller for online solutions. 6) It includes map encoding, which provides dimensionality reduction and the ability to work with different maps. The following works seem to be the closest to ours concerning these properties. \parencite{kur22} satisfies 1), 2), and 4). However, it learns a trajectory for the single map offline. \parencite{med23} satisfies 3), 5), and 6) however, it is intended for the different tasks. \parencite{ada22, abd23} satisfy 1), 2), 3), and 5), however, they process vision data instead of obstacle map and work with simple-shaped obstacles. Other works satisfy less number of properties.

Unique property 7) of our approach is encoding the footprint of the mobile robot. It allows one to use a single model for mobile robots with different shapes. Note that in this work, we only prove a concept of footprint encoding: our training set consists of samples with two various footprints, and we show that the networks learn their collision model. We consider the deeper study of footprint learning (including footprint generalization) to be a part of the future work.

\section{Control approach}
In this section, we discuss a model predictive controller, which is used for local planning. Neural networks are considered to be black boxes, which take inputs and provide outputs within the control architecture. Their internal content is discussed in a further section. We first describe the formal statement of a local optimization problem and then discuss the controller that solves this statement.

\subsection{MPC Statement}

Local trajectory optimization may be formulated as a nonlinear model predictive control problem with continuous dynamics and discrete control:
\begin{subequations} \label{mpc_d}
\begin{equation} \label{mpc_d_1}
\{\mathbf{x}_{opt}[i],\mathbf{u}_{opt}[i]\}_{i=k}^{k+p}=arg\min\sum_{i=k}^{k+p}J(\mathbf{x}[i],\mathbf{u}[i],\mathbf{p}[i]),
\end{equation}
s.t.
\begin{equation} \label{mpc_d_2}
\begin{aligned}
\frac{dx_1[i]}{dt}=f_1(\mathbf{x}[i],\mathbf{u}[i],\mathbf{p}[i]),\\
\frac{dx_2[i]}{dt}=f_2(\mathbf{x}[i],\mathbf{u}[i],\mathbf{p}[i]),\\
\dots\\
\frac{dx_\mu[i]}{dt}=f_\mu(\mathbf{x}[i],\mathbf{u}[i],\mathbf{p}[i]),
\end{aligned}
\end{equation}
\begin{equation} \label{mpc_d_4}
\begin{aligned}
h_1(\mathbf{x}[i],\mathbf{u}[i],\mathbf{p}[i])\leq0,\\
h_2(\mathbf{x}[i],\mathbf{u}[i],\mathbf{p}[i])\leq0,\\
\dots\\
h_\chi(\mathbf{x}[i],\mathbf{u}[i],\mathbf{p}[i])\leq0.\\
\end{aligned}
\end{equation}
\end{subequations}

Here $p$ is prediction horizon, $\mathbf{x}[i]$ is $\mu$-size state vector (at the beginning of step $[i]$), $\mathbf{u}[i]$ is $\nu$-size control vector (constant within step $i$) $\mathbf{p}[i]$ is $\kappa$-vector of process parameters (relevant for the step $i$). \eqref{mpc_d_1} specify the cost function $J$: a sum of functions $J[i]$, which are calculated for each node.  \eqref{mpc_d_2} define continuous dynamics of the process. \eqref{mpc_d_4} is a set of inequality constraints which must be satisfied within the whole process. Optimization procedure aims to find the reference of $\{\mathbf{x}_{opt}[i],\mathbf{u}_{opt}[i]\}_{i=k}^{k+p}$ that provide minimum $J$. Note that in this work we use this statement for defining local trajectory of the robot (i.e. $\mathbf{x}_{opt}[i]_{i=k}^{k+m}$). Control of trajectory execution may be either provided by other control method or achieved by direct execution $u_{opt}$.

The view of the equation \eqref{mpc_d_2} depend on the construction of the mobile robot. In this work we consider two different models: a differential drive model (relevant for our real-robot experiments) and a bicycle model (used in numerical experiments). Differential drive model is specified as follows:
\begin{equation}\label{diff_drive_model}
\begin{aligned}
\frac{dx}{dt} = v\cos\theta, \\ 
\frac{dy}{dt} = v\sin\theta, \\ 
\frac{dv}{dt} = a, \\ 
\frac{d\theta}{dt} = \omega.
\end{aligned}
\end{equation}

State vector $\mathbf{x} = (x,y,v,\theta)^T$ include cartesian position of the robot $x,y$ , its linear velocity $v$, and its orientation $\theta$. Control vector $\mathbf{u} = (a,\omega)^T$ include linear acceleration $a$ and angular velocity $\omega$. 

For bicycle model $\mathbf{u} = (a,\delta)^T$ where $\delta$ is steering angle. .

\begin{equation}\label{bicycle_model}
\begin{aligned}
\frac{dx}{dt} = v\cos\theta, \\ 
\frac{dy}{dt} = v\sin\theta, \\ 
\frac{dv}{dt} = a, \\ 
\frac{d\theta}{dt} = \frac{v}{L}\tan\delta.
\end{aligned}
\end{equation}

For the optimal control problem of this model, the following cost function is introduced: 
\begin{equation} \label{cost}
\begin{aligned}
J[i] = J_s(\mathbf{x}[i],\mathbf{u}[i],\mathbf{x}_{r}[i]) + J_o(\mathbf{x}[i],\mathbf{p}_o[i]). \\
\end{aligned}
\end{equation}

$J_s$ term enforce the trajectory to follow the reference values $\mathbf{x}_{r}$ from the global plan, while $J_o$ term push the trajectory farther from obstacles. $\mathbf{p}_o[i]$ is a vector of obstacle-related parameters. Whole parameter vector for the system \eqref{mpc_d} is $\mathbf{p}[i] = ((\mathbf{x}_r[i])^T,(\mathbf{p}_o[i])^T)^T$. In our approach, the neural network is applied to compute $J_o$ while $J_s$ is calculated as follows:
\begin{equation} \label{cost_state}
J_s[i] =  \sum_{j=1}^\mu w_{xj}(x_j[i]-x_{j(ref)}[i])^2 + \sum_{k=1}^\nu w_{uk}u_k^2[i]  \, 
\end{equation}

Here $w_{xj}, w_{uk}$ are the weights of the respective terms, $x_{j(ref)}[i]$ is a reference value of the respective state (taken from the global plan). 

Constraint-based trajectory optimizers like CIAO \parencite{sch20a, sch20b} use equation \eqref{mpc_d_4} to provide collision avoidance. Contrary, we express collision avoidance in \eqref{mpc_d_1} and use \eqref{mpc_d_4} only for box constraints of the separate variables.

\subsection{Control architecture}
\begin{figure} [t]
\centering
\includegraphics[width=0.48\textwidth]{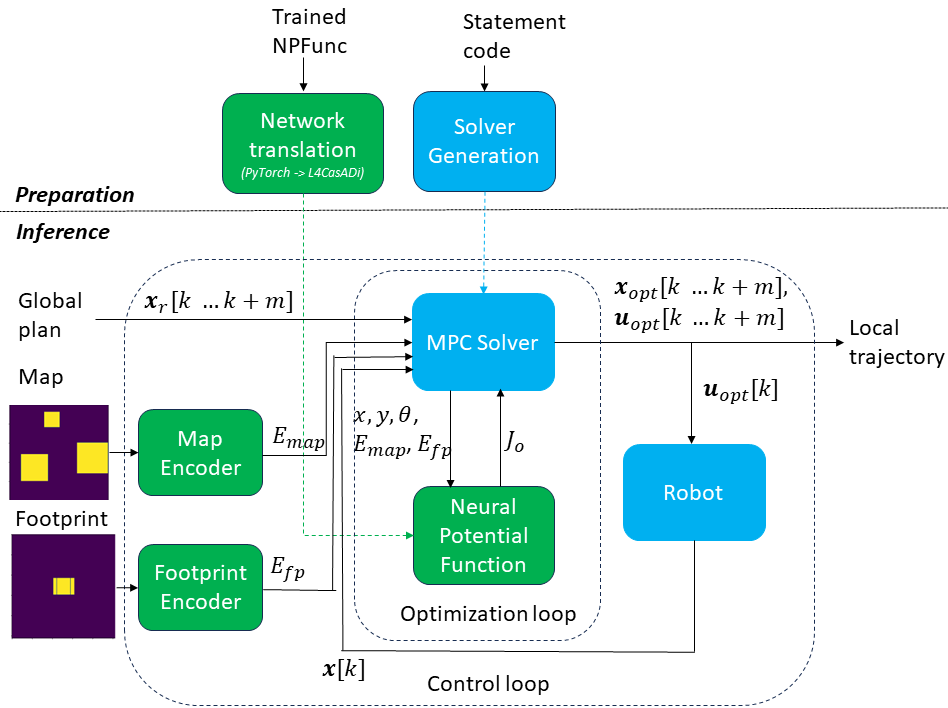}
\caption{Proposed architecture of the controller. Image encoders work outside the optimization loop and we just need to run them once before each optimization procedure. NPFunction works within the optimization loop and the solver uses its gradients to find a safer trajectory.}
\label{fig:npfield}
\end{figure}

An architecture of our controller is shown in Fig. \ref{fig:npfield}, which is a more detailed version of Fig. \ref{fig:visual_abstract}. Solution of the problem \eqref{mpc_d} is obtained iteratively using Sequential Quadratic Programming (\textit{optimization loop} in Fig. \ref{fig:npfield}). MPC controller uses the solution to update the trajectory online (\textit{control loop} in Fig. \ref{fig:npfield}). At the timestep $k$ it optimizes the trajectory for the next $p$ steps, and then the optimized control inputs are sent to the robot for the next $m$ steps (i.e. $m$ is the control horizon). After that optimization is repeated for the steps from $k+m$ to $k+m+p$.

MPC-solver is intended to provide solution of the problem \eqref{mpc_d} with \eqref{cost_state} as \eqref{mpc_d_1} and \eqref{diff_drive_model} as \eqref{mpc_d_2}. During optimization, it communicates with integrated NPFunction, which provides values and gradients of $J_{obst}$. 

The objective of the neural network is to project the robot’s footprint, obstacle map, and robot poses onto a differentiable obstacle-repulsive potential surface. Consequently, for each coordinate within the range of the map, the neural network outputs a corresponding potential value. 
To ensure computational feasibility, we partitioned the neural network into two blocks: a map and footprint encoder, and a final coordinate potential predictor. Encoders compress high-dimensional maps into a compact representation, 
thereby enabling the computation of Jacobian and Hessian matrices of the control problem within the solver. Encoders work outside the optimization loop: provided embeddings $E_{map}$ and $E_{fp}$ are sent to the solver as obstacle-related problem parameters $\mathbf{p}_o$. This means that we assume the local map and robot footprint to be fixed within the prediction horizon. While the robot is following the global plan, the local map slides according to its current position and actual sensor data.

\section{NEURAL POTENTIAL FIELD}
In this section we discuss our neural model for calculating obstacle repulsive potential $J_o$. First, we briefly describe architecture of our network, then we introduce our strategy for generating the training set.

\subsection{Network architecture}

\begin{figure} [t]
\centering
\includegraphics[width=0.48\textwidth]{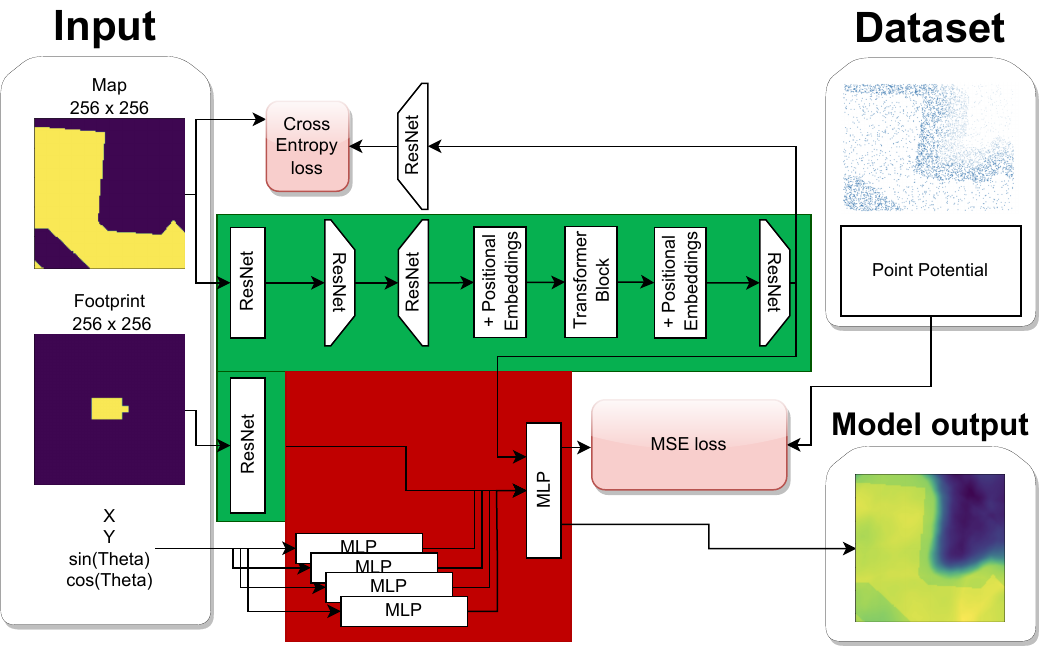}
\caption{Proposed architecture of the neural network. The green component represents the robot footprint and map encoder, which generates embeddings that are consistent for all coordinates within the map. The red component signifies the final coordinate potential predictor, which contains an order of magnitude fewer parameters.}
\label{fig:net_arc}
\end{figure} 

Proposed neural architecture is shown in Fig. \ref{fig:net_arc}. It consists of three primary components: a ResNet Encoder, a Spatial Transformer, and a ResNet Decoder. ResNet blocks are used with the objective of extracting local features from the obstacle map which contains a lot of corners and narrow passages. The Spatial Transformer utilizes the self-attention mechanism \cite{attention} to establish global relations among these features, assessing the significance of one feature in relation to others. Consequently, we employ the positional embedding technique from Visual Transformers \cite{vit}. Lastly, the ResNet Decoder processes the transformed feature maps to generate the final output.

To mitigate the model’s tendency to truncate critical details of the obstacle map necessary for navigation, we incorporated a map reconstruction loss based on Cross-Entropy. For the predictions of potential points, we employed the Mean Squared Error (MSE) loss. 

\subsection{Training data}
One unit of the training set include $\{I_{map}, I_{fp}, x, y, \theta, J_o\}$, where $I_{map}$ and $I_{fp}$ are 2D images of the obstacle map and the robot footprint. The dataset should include various samples of robot positions from various maps. The maps are cropped from the MovingAI planning dataset \parencite{stu12}. For each map, we generate a set of random robot poses and calculate reference values for them using the following algorithm.
\begin{enumerate}
    \item Obstacle map is transformed into a costmap: 
    \begin{enumerate}
        \item Signed distance function (SDF) is calculated algorithmically for each cell on the map. SDF is equal to the distance from the current cell to the nearest obstacle border. It is positive for free space cells and negative for obstacle cells.
        \item Repulsive potential is calculated for each cell:
       $J_o =  w_1(\pi / 2 + arctan(w_2-w_2SDF))$. This is a sigmoid function, which is low far from obstacles, asymptotically strives to $w_1$ inside obstacles, and has maximum derivative on the obstacle border.        
    \end{enumerate}
    \item Collision potential is calculated for each random pose of the robot within the submap. For this purpose robot's footprint is projected onto the map according to the pose. The maximum potential among the footprint-covered cells is chosen as a collision potential. As an alternative approach, we tried to compute collision potential as an integral potential over footprint. This trial did not provide learning of the useful potential function.
\end{enumerate}

\begin{figure} [t]
\centering
\includegraphics[width=0.3\textwidth]{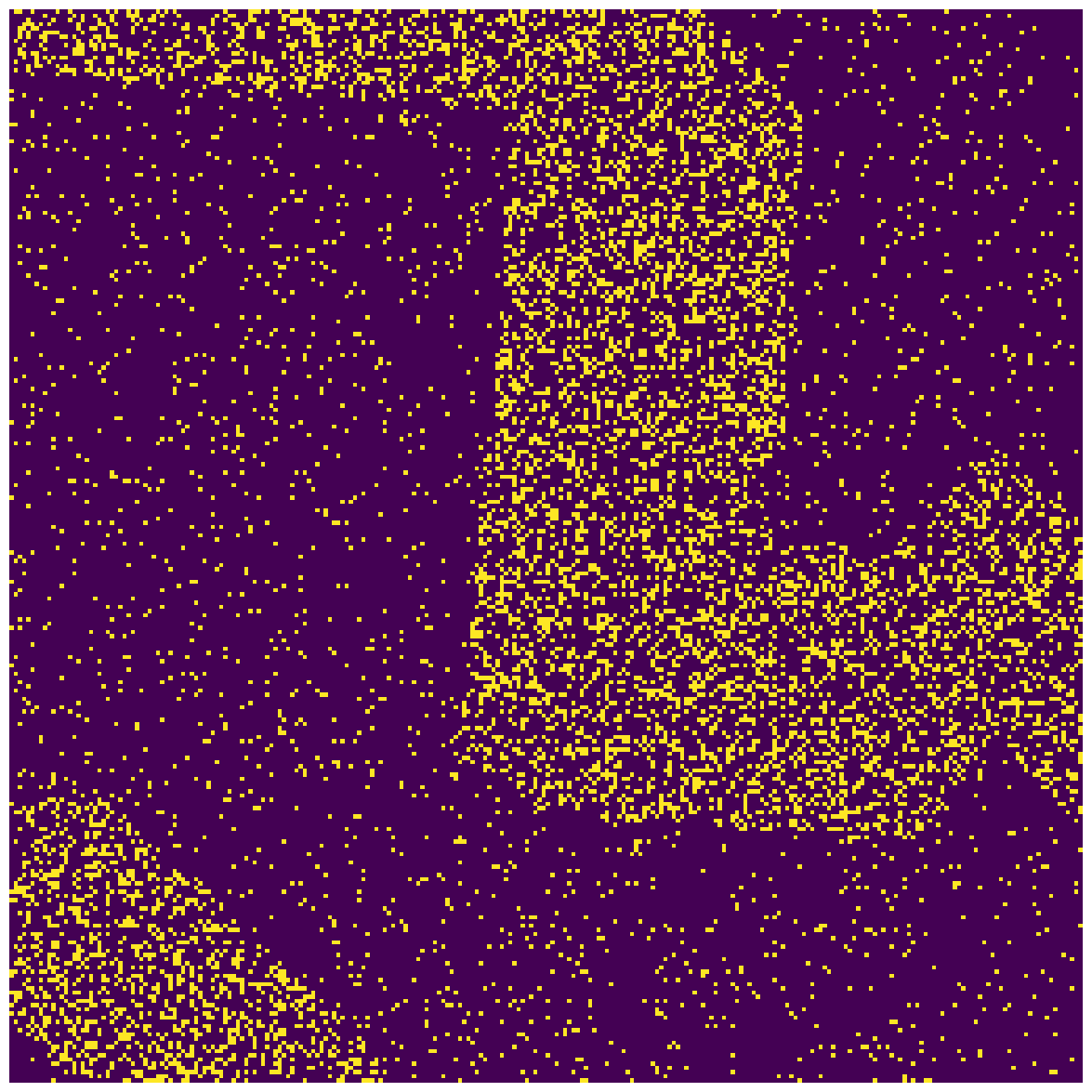}
\caption{Example for distribution of samples in a map. }
\label{fig:sampling_points}
\end{figure}

A pivotal aspect of the training process was the dataset sampling strategy. Utilizing a random sampling strategy across the map led to the network overfitting to larger values and disregarding narrow passages. This is because of the walls, which are statistically overwhelming compared to free space, but are irrelevant for navigation as we explicitly avoid planning through obstacles. To address this, we modified the sampling strategy such that 80\% of points are sampled with intermediate potential values. The figure \ref{fig:sampling_points} shows the distribution of point samples in a map. The area with a high density of points represents the area surrounding and close to obstacles, while obstacles have a little effect on the path of movement in the areas with a low density of points.

\section{Implementation}
We consider nonlinear MPC task statement, which may be solved via Interior Point (IP) or Sequential Quadratic Programming (SQP). Modern frameworks provide the possibility for realtime execution of these methods. IPOPT \parencite{IPOPT} and ForcesPro \parencite{zan17} implement IP, while ACADO \parencite{hou11a,hou11b} and Acados \parencite{acados} implement SQP. These frameworks rely on a more low-level CasADi framework \parencite{and19} for algorithmic differentiation. We implement our MPC solver with Acados framework, is which is the newest one and provide the fastest execution. 

Use of deep neural network within Acados solver require the specific integration tool. Two libraries are relevant for this task: ML-CasADi \parencite{ml_casadi} and L4CasADi \parencite{l4casadi}. Both provide the CasADi description of Pytorch \parencite{pas19} neural models. However, the first method was proposed and used for replacing complex models with local Taylor approximations to enable real-time optimization procedures, while the second method provides a complete mathematical description of the Pytorch model by CasADi formula. For our Pytorch model which describes the neural potential field of the obstacles surrounding the path of the robot, L4CasADi is more suitable because the description of the whole model is needed and not only at a linearizing point.
ML provide lightweight local approximation of the complex neural model. This approximation constrain the use of ML-CasADi for the functions with complicated landscapes. Novel framework L4CasADi  do not use such approximations. It provides integration of the deep neural models into real-time CasADi-based optimization. We use the L4CasADi to provide optimization over NPFunction. To our knowledge, our work is the first one, which exploits L4CasADi for neural cost terms instead of the neural dynamic model. 

Our local planner works together with Theta* \cite{nas07} global planner, which generates global plans as polylines. Note that Theta* uses a simplified version of the robot footprint (a circle with a diameter equal to the robot width) as it fails to provide a safe path with a complete footprint model. 
This simplified model does not guarantee the safety of the global plan, therefore the safety of the trajectory is provided by our local planner.

We consider obstacle maps to have a 256×256 resolution, where each pixel corresponds to 2×2 centimeters of the real environments (i.e. size of the map is 5.12×5.12 meters). We collected a dataset based on the MovingAI \parencite{stu12} city maps. It includes 4,000,000 samples taken from 200 maps with 2 footprints. Both footprints correspond to a real Husky UGV mobile manipulator with an UR5 robotic arm. The first one is with a folded arm, the second one is with an outstretched arm. 10,000 random poses of the robot were generated for each map with each footprint. Weighting coefficients for reference potential were set to $w_1 = 15$ and $w_2 = 10$, while the prediction horizon was set to $p=30$. Dataset generation took 40 hours on the Intel Core i5-10400F CPU.

Our neural network consists of 5 million parameters, with 500,000 allocated to ResNet encoders. Encoders project each (256×256) map and robot footprint into (1×4352) embeddings. The robot’s pose, represented as $X,Y,sin(\theta),cos(\theta)$, is transformed into (1,32) embeddings. The model was trained over a span of 24 hours on a server equipped with a single Nvidia Tesla V100 card with 32GB of memory.

\section{Experiments}
In this section we first present numerical comparison of our approach with other planning methods. Then we discuss effects of varying some hypreparameters of our method. Finally we show the experiments on a real robot.

\subsection{Comparative studies}

\begin{figure} [t]
\centering
\includegraphics[width=0.48\textwidth]{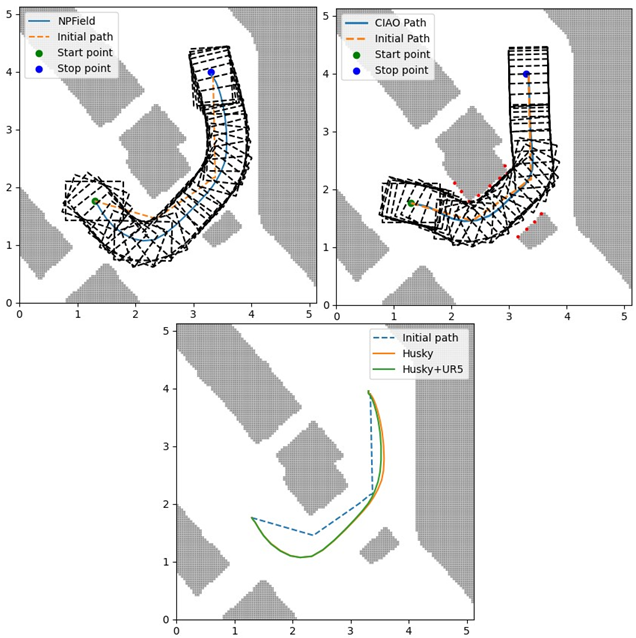}
\caption{Example scenario for local planning. Left: NPField trajectory. Right: CIAO trajectory. Bottom: trajectory curves for different footprints. }
\label{fig:footprints}
\end{figure} 
All experiments reported in this subsection were 1) conducted with the bicycle model of process dynamics, and 2) conducted on the maps from the MovingAI dataset \parencite{stu12}, which were not used for network training.
\subsubsection{Illustrative example and comparison with trajectory optimizers}
An example of the trajectory generated with our planner is shown in Fig. \ref{fig:footprints} on the left. A global plan in the form of a polyline is turned into a smooth and safe trajectory. Initially, the robot turns from the obstacle and deviates from the global path, then smoothly returns to it, reaching the goal position.

As a proof of concept for footprint encoding, we provide the following experiment. Consider the global plan, where the robot first moves towards the flat wall, then turns and moves parallel to the wall. In this case, a robot with a folded arm turns a little later than the one with a folded arm. Such a behavior may be seen in Fig.  \ref{fig:footprints}, bottom. The yellow curve relates to the outstretched arm, while the green curve relates to the folded arm.
This behavior shows that the model learns different properties of two footprints, which are useful for safer trajectory planning.

We compared NPField trajectories with CIAO \cite{sch20a}  trajectory optimizer, which is based on convex approximation of the free space around the robot. The CIAO-generated trajectory is shown in Fig. \ref{fig:footprints} on the right. It may be seen that it keeps the robot near obstacles, nearly touching their edges. When testing on more diverse scenarios, CIAO could not find the feasible path in nearly half of the cases. It may be connected with the fact that CIAO implements collision avoidance as a set of inequality constraints, which are not differentiated during optimization. Therefore, it only checks the fact of the collision and does not balance between safety and path deviation in the cost function. 

\subsubsection{Comparison on BenchMR}
We compare our algorithm with the baselines on 20 scenarios using BenchMR \parencite{hei21} framework. The tasks include moving through the narrow passages similar to those shown in Fig. \ref{fig:footprints}. We compare standard metrics: planning time, path length, smoothness, and angle-over-length (for all, lower value is better). We also introduce our custom metric, "safety distance" (minimum value of the SDF). 

\begin{table}[t]
\centering
\begin{tabular}{||m{1.4cm} |m{0.9cm}|m{0.9cm}| m{0.9cm} | m{0.9cm} | m{1cm} ||} 
 \hline
  Planner & Time, s  & Length, m & Smooth-ness & AOL & Safety distance, m \\ [0.5ex] 
 \hline\hline
 RRT* & 11  & 2.27  & 0.008 & 0.005  & 0.048 \\
 \hline
RRT & 0.013  & 2.72  & 0.012 & 0.010  & 0.148 \\
 \hline
InformedRRT & 11  & 2.27  & 0.006 & 0.004  & 0.041 \\
 \hline
 SBL & 0.062  & 4.99  & 0.055 & 0.042  & 0.049 \\
 \hline
RRT+GRIPS & 0.013  & 2.44  & 0.009 & 0.004  & 0.151 \\
  \hline
$\theta*$+NPField (ours)& 0.063  & 2.33  & 0.002 & 0.006  & 0.116 \\
 \hline
\end{tabular}
\caption{Comparative studies on BenchMR scenarios}
\label{tab:comparative}
\end{table}

The results are given in table \ref{tab:comparative}. We compare our stack (Theta* + NPField) with state of the art planners: RRT \parencite{lav01}, RRT* \parencite{kar11}, Informed RRT \parencite{gam14}, SBL \parencite{hsu97} and RRT with GRIPS \parencite{hei18} smoothing. We do not provide the results for  PRM \parencite{kav96}, PRM* \parencite{kar11}, BIT* \parencite{gam20}, KPIECE1 \parencite{suc09}, Theta* with CIAO \parencite{sch20a} optimization, Theta* with CHOMP \parencite{sch14} optimization as they were able to generate a successful plan for less than a half of tasks. This result is particularly important for Theta* + CIAO and Theta* + CHOMP, as they are optimization-based planers similar to our approach and use the same global plans. However, they could not handle considered scenarios due to collisions (CHOMP) or failure to find a result (CIAO). Results in the table show that our stack is generally comparable to other planners. It provides nearly the shortest path length, the best smoothness, a good AOL, and a good safety distance. 

\begin{figure} [t]
\centering
\includegraphics[width=0.24\textwidth]{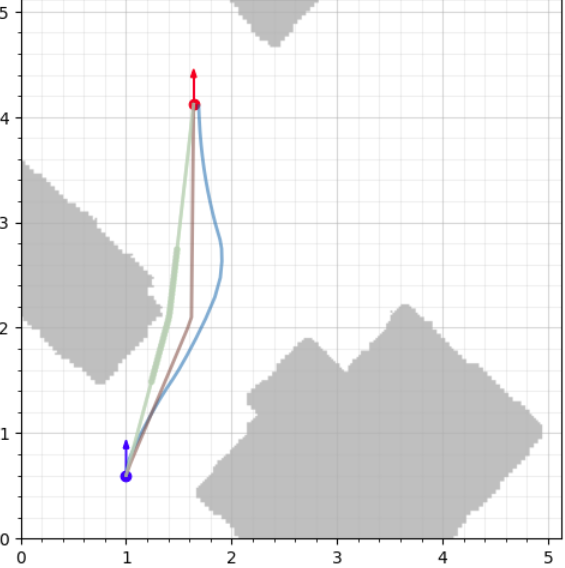}
\caption{Example comparison of NPField (blue curve) and CHOMP (green curve). CHOMP converges to the shortest path ignoring obstacle danger.}
\label{fig:footprints}
\end{figure} 

Computation time has the same order of magnitude with the fastest methods. We cannot specify an approach, which is definitely better than ours (RRT with GRIPS is fast and safe but provides less smooth trajectories). Performance measurements were made on Intel Core i5-10400F CPU. Note that Acados solver need to warmup before reatime use: first execution of the optimization procedure may take about one second; after that the solver work faster. One optimization take 60-70 ms, where data encoding take around 10 ms, while Acados solution take the rest 50-60 ms.

\subsection{Ablation studies}
We also compare various versions of our algorithm on the same set of scenarios. These versions are different each other by the weights of the potential function which is used to calculate the potentials in training dataset, the features used for training the model, or the distribution the sampling points in the training maps.

\begin{figure} [t]
\centering
\includegraphics[width=0.3\textwidth]{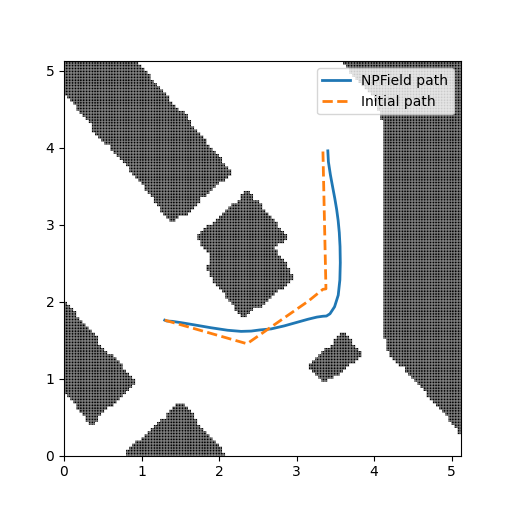}
\caption{Example of path with integral potential}
\label{fig:integral_potential}
\end{figure}

First, we consider the situation when the reference collision potential (see subsection IV.B) is calculated as an integral value over footprint instead of choosing the maximum value. We had a hypothesis that such an approach could lead to better learning of the relative geometry of the object. However, it did not lead to an useful model. In our experiments the network provides incorrect  results systematically (example is given in Fig. \ref{fig:integral_potential}).

\begin{figure} [t]
\centering
\includegraphics[width=0.3\textwidth]{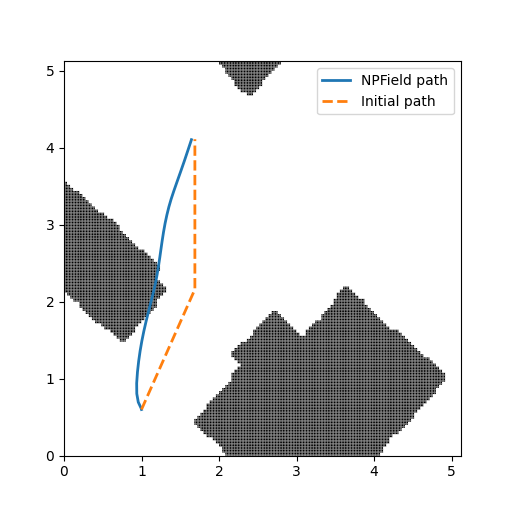}
\caption{Example of path with different weights model}
\label{fig:alt_weights}
\end{figure}

Similar incorrect results were measured for an alternative choice of weighting coefficient $w_2$ for calculating the training values of the repulsive potential. The idea was to make the potential landscape more gentle and provide better optimization from incorrect initial guess. In practice it lead to bad learning of the map properties, see example rezult for $w_2 = 1$ in Fig. \ref{fig:alt_weights}.

\begin{figure} [t]
\centering
\includegraphics[width=0.48\textwidth]{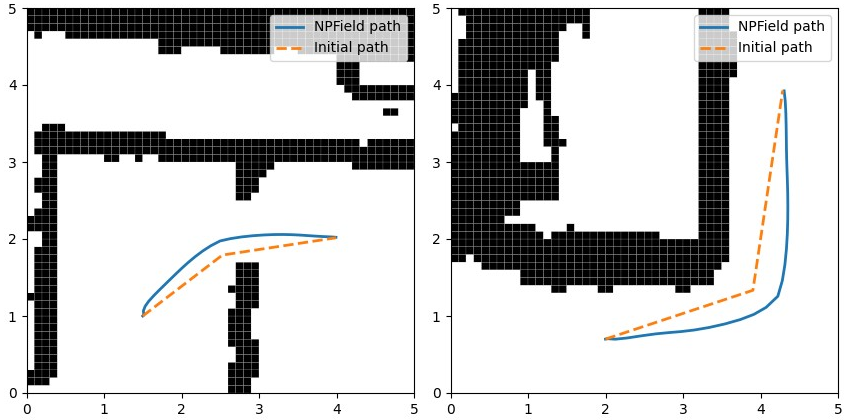}
\caption{Example of paths on a low-resolution map.}
\label{fig:low_resolution}
\end{figure}

Our third ablation experiment was connected with the varying resolution of the obstacle map and robot footprint. We consider the situation when one cell of the grid correspond to 10x10 cm instead of 2x2 cm. These two sets are specified by a practical resolution of the global map and the local map respectively. The global map of the environment is stored in the memory of the robot, while the local map include actual data from the sensors. The size of the submap is 50x50 pixels (i.e. 5x5 meters). This size allow us to reduce the complexity of the neural network: total number of parameters is 1.4M instead of 5M; the size of the map embedding is 1161 instead of 4352. This archintucture still allow correct solution of the planning task; examples are provided in Fig. \ref{fig:low_resolution}. Surprisingly reducing the model complexity did not affect the performance of the solver: it still take around 70 ms to make an optimization.

\subsection{Real Robot Experiments}
\begin{figure} [t]
\centering
\includegraphics[width=0.48\textwidth]{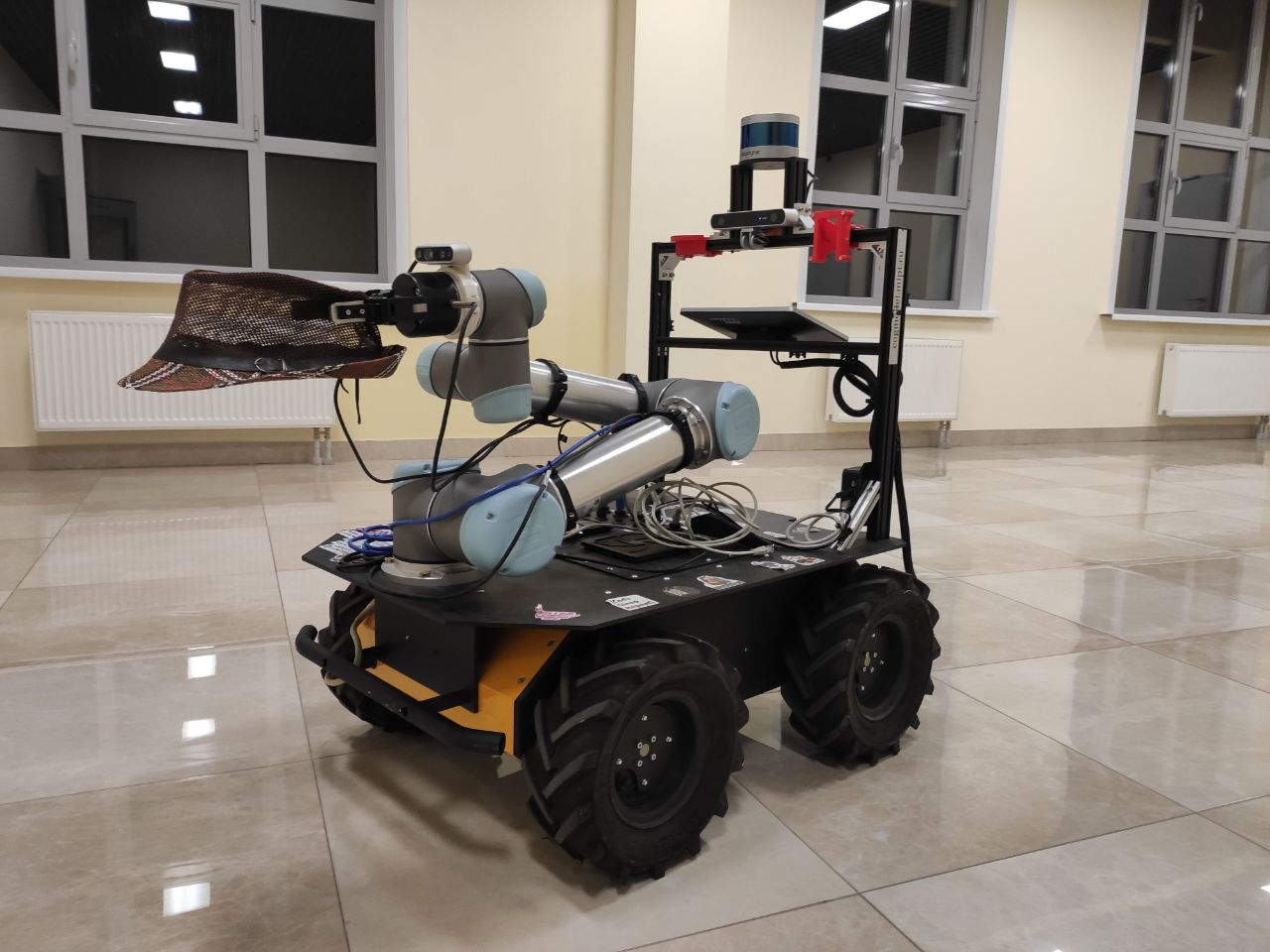}
\caption{Husky mobile manipulator transforming a hat. A footprint with an outstretched manipulator is valid for this task.}
\label{fig:real_robot}
\end{figure}

We deploy our approach on a real Husky UGV mobile manipulator as a ROS module for MPC local planning and control, which works with Theta* global planner. The testing scenario includes hat transportation through a twisty corridor. The manipulator is holding the hat in an outstretched configuration (see Fig. \ref{fig:real_robot}). Acados optimizer run Intel Core i5-10400F CPU and communicate with the robot in real time as a remote ROS-node with control horizon equal to one step. Therefore, a more complicated concave footprint is valid. Scenario execution may be seen in the accompanying video (see \url{https://github.com/cog-isa/NPField}).

\section{CONCLUSIONS}
We propose a novel approach to local trajectory planning, where a Model Predictive Controller uses the neural model to estimate collision danger as a differentiable function. Our NPField neural architecture consists of encoders and NPFunction blocks. Encoders provide a compact representation of the obstacle map and robot footprint; this compact representation is sent to the MPC solver as a vector of problem parameters. NPFunction is integrated into the optimization loop, and its gradients are used for trajectory correction. We implement our controller using Acados MPC framework and L4CasADi tool for integrating deep neural models into MPC loop. Our approach allows the robot with a complicated footprint to successfully navigate among the obstacles in real time. A planning stack Theta* + NPField showed comparable results with sample-based planners on the BenchMR testing framework. The code for our approach is presented at \url{https://github.com/cog-isa/NPField}.

We consider our work a starting point for further research on neural potential estimation for kinodynamic planning for various robotic systems in various environments. Trajectory planning on more complex maps (e.g. elevation maps) is a promising topic of the future research. Another important aspect is further research on footprint encoding, which may be useful for planning the trajectories of the robotic system with changing footprints (e.g. mobile manipulators under whole-body control).








\printbibliography

\end{document}